\documentclass[%
 amsmath,amssymb,
 aps,
 prl,
 twocolumn,
]{revtex4-2}

\usepackage{color}
\usepackage{graphicx}
\usepackage{dcolumn}
\usepackage{bm}

\renewcommand{\arraystretch}{0.75}

\begin{document}

\preprint{APS/123-QED}

\title{Reservoir computing with diverse timescales for prediction of multiscale dynamics}

\author{Gouhei Tanaka$^{1,2,3,*}$}
 \author{Tadayoshi Matsumori$^4$}%
 \author{Hiroaki Yoshida$^4$}
 \author{Kazuyuki Aihara$^1$}
\affiliation{%
  $^1$International Research Center for Neurointelligence, The University of Tokyo, Tokyo 113-0033, Japan \\
  $^2$Department of Electrical Engineering and Information Systems, Graduate School of Engineering, The University of Tokyo, Tokyo 113-8656, Japan. \\
  $^3$Department of Mathematical Informatics, Graduate School of Information Science and Technology, The University of Tokyo, Tokyo 113-8656, Japan.\\
  $^4$Toyota Central Research and Development Laboratory Inc. Tokyo 112-0004, Japan.
}%




\date{\today}

\begin{abstract}
  Machine learning approaches have recently been leveraged as a substitute or an aid for physical/mathematical modeling approaches to dynamical systems. To develop an efficient machine learning method dedicated to modeling and prediction of multiscale dynamics, we propose a reservoir computing (RC) model with diverse timescales by using a recurrent network of heterogeneous leaky integrator (LI) neurons. We evaluate computational performance of the proposed model in two time series prediction tasks related to four chaotic fast-slow dynamical systems. In a one-step-ahead prediction task where input data are provided only from the fast subsystem, we show that the proposed model yields better performance than the standard RC model with identical LI neurons. Our analysis reveals that the timescale required for producing each component of target multiscale dynamics is appropriately and flexibly selected from the reservoir dynamics by model training. In a long-term prediction task, we demonstrate that a closed-loop version of the proposed model can achieve longer-term predictions compared to the counterpart with identical LI neurons depending on the hyperparameter setting.
\end{abstract}

\maketitle



{\it Introduction.--} Hierarchical structures composed of macroscale and microscale components are ubiquitous in physical, biological, medical, and engineering systems \cite{vlachos2005review,matouvs2017review,deisboeck2011multiscale,walpole2013multiscale}. Complex interactions between such diverse components often bring about multiscale behavior with different spatial and temporal scales. To understand such complex systems, {\it multiscale modeling} has been one of the major challenges in science and technology \cite{weinan2011principles}. An effective approach is to combine established physical models at different scales by considering their interactions. However, even a physical model focusing on one scale is often not available when the rules (e.g. physical laws) governing the target system are not fully known. In such a case, another potential approach is to employ a machine learning model fully or partly \cite{pathak2018model,pathak2018hybrid}. It is a challenging issue to integrate machine learning and multiscale modeling for dealing with large datasets from different sources and different levels of resolution \cite{alber2019integrating}. To this end, it is significant to develop robust predictive machine learning models specialized for multiscale dynamics.

We focus on a machine learning framework called {\it reservoir computing} (RC), which has been mainly applied to temporal pattern recognition such as system identification, time series prediction, time series classification, and anomaly detection \cite{jaeger2001echo,jaeger2004harnessing,maass2002real,verstraeten2007experimental,lukovsevivcius2009reservoir,pathak2018model}. The echo state network (ESN) \cite{jaeger2001echo,jaeger2004harnessing}, which is a major RC model, uses a recurrent neural network (RNN) to nonlinearly transform an input times series into a high-dimensional dynamic state and reads out desired characteristics from the dynamic state. The RNN with fixed random connection weights serves as a {\it reservoir} to generate an echo of the past inputs. Compared to other RNNs where all the connection weights are trained with gradient-based learning rules \cite{werbos1990backpropagation}, the ESN can be trained with much lower computational cost by optimizing only the readout parameters with a simple learning method \cite{lukovsevivcius2012practical}.

The timescale of reservoir dynamics in the original ESN \cite{jaeger2001echo} is almost determined by that of input time series. However, the timescale of a desired output time series is often largely different from that of the input one depending on a learning task. Therefore, the ESN with leaky integrator neurons (LI-ESN) has been widely used as a standard model to accommodate the model output to temporal characteristics of the target dynamics \cite{jaeger2007optimization}. The LI-ESN has a leak rate parameter controlling the update speed of the neuronal states in the reservoir. For multi-timescale dynamics, it is an option to use the hierarchical ESN combining multiple reservoirs with different timescales \cite{jaeger2007discovering,manneschi2021exploiting}, where the leak rate is common to all the neurons in each reservoir but can be different from one reservoir to another. In the above models, the leak rate in each reservoir is set at an optimal value through a grid search or a gradient-based optimization.

In contrast to the above approach, here we propose an ESN with diverse timescales (DTS-ESN), where the leak rates of the LI neurons are distributed. Our aim is to generate reservoir dynamics with a rich variety of timescales so as to accommodate a wide range of desired time series with different timescales. The main advantage of the DTS-ESN is flexible adjustability to multiscale dynamics. Moreover, the idea behind the proposed model opens up a new possibility to leverage heterogeneous network components for network-type physical reservoirs in implementation of RC hardware \cite{tanaka2019recent} as well as for other variants of ESNs and RC models \cite{gallicchio2017deep,tamura2021transfer,li2022multi,akiyama2022computational}. Our idea is also motivated by the role of heterogeneity of biological neurons in information flow and processing \cite{di2021optimal}.

Some previous works considered heterogeneity of system components in RC systems. A positive effect of heterogeneous nonlinear activation functions of reservoir neurons on prediction performance is reported \cite{tanaka2016exploiting}, but its explicit relation to timescales of reservoir states is unclear. In a time-delay-based reservoir \cite{appeltant2011information} where the timescales of system dynamics are mainly governed by the delay time and the clock cycle, a mismatch between them can enhance computational performance \cite{paquot2012optoelectronic,stelzer2020performance}. In a time-delay-based physical reservoir implemented with an optical fiber-ring cavity, it was revealed that the nonlinearity of the fiber waveguide is essential for high computational performance rather than the nonlinearity in the input and readout parts \cite{pauwels2019distributed}. Compared to these time-delay-based reservoirs with a few controllable timescale-related parameters, our model based on a network-type reservoir can realize various timescale distributions by setting different heterogeneity of leak rates.

\begin{figure}[t]
\includegraphics[width=0.65\linewidth]{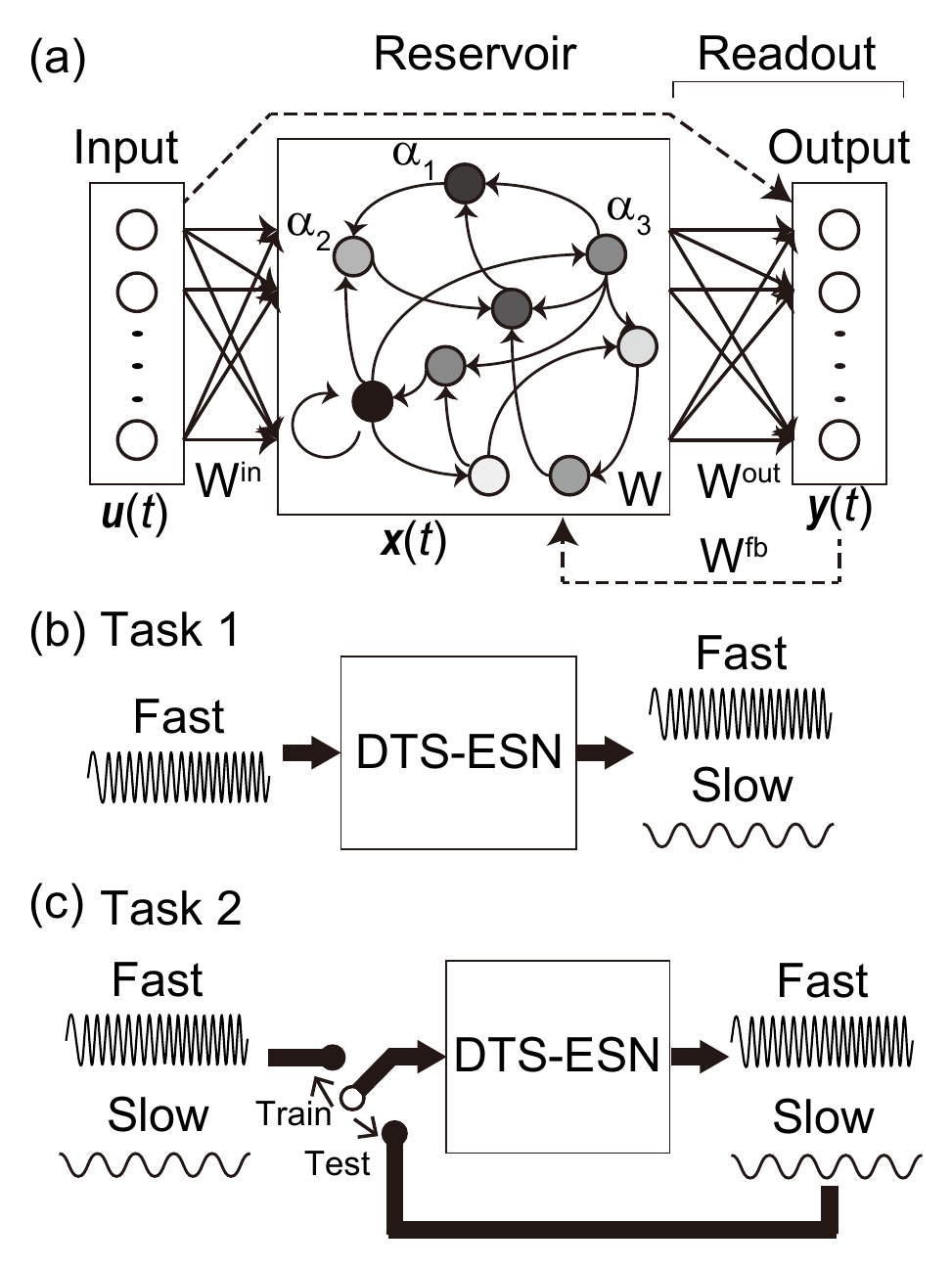}
\caption{\label{fig:DTSESN} (a) A schematic of the DTS-ESN. An input time series $\mathbf{u}(t)$ given to the input layer is transformed into a reservoir state $\mathbf{x}(t)$, and then processed in the readout to produce an output time series $\mathbf{y}(t)$. Only $W^{\rm out}$ is trainable. The dashed arrows indicate optional connections. The reservoir neuron $i$ has leak rate $\alpha_i \in [\alpha_{\rm min}, \alpha_{\rm max}]$. (b) Task 1 where the whole fast-slow dynamics is predicted only from the fast dynamics using an open-loop model. (c) Task 2 where the whole fast-slow dynamics is predicted with a closed-loop model.}
\end{figure}

{\it Methods.--} The DTS-ESN consists of a reservoir with heterogeneous LI neurons and a linear readout as illustrated in Fig.~\ref{fig:DTSESN}(a). The DTS-ESN receives an input time series $\mathbf{u}(t)$ in the input layer, transforms it into a high-dimensional reservoir state $\mathbf{x}(t)$, and produces an output time series $\mathbf{y}(t)$ as a linear combination of the states of the reservoir neurons. With distributed leak rates of reservoir neurons, the DTS-ESN extends greatly the range of timescales that can be realized by the LI-ESN \cite{jaeger2007optimization}.

In this study, the capability of the DTS-ESN is evaluated in two types of time series prediction tasks related to chaotic fast-slow dynamical systems. One is a one-step-ahead prediction task (Task 1) where input data are given only from the fast subsystem as depicted in Fig.~\ref{fig:DTSESN}(b). An inference of hidden slow dynamics from observational data of the fast dynamics is challenging but beneficial in reducing the effort involved in data measurement. An example in climate science is to predict slowly changing behavior in the deep ocean such as the temperature. The behavior is known to give an important feedback to the fast dynamical behavior of the atmosphere, land, and near-surface oceans~\cite{seshadri2017fast}, but the observation in the deep ocean remains as a major challenge~\cite{meinen2020observed}.
The other is an autoregressive prediction task (Task 2) where a closed-loop version of the proposed model is used for a long-term prediction in a testing phase as illustrated in Fig.~\ref{fig:DTSESN}(c). RC approaches have shown a strong potential in long-term predictions of chaotic behavior \cite{pathak2018model,pathak2018hybrid}. We examine the effect of distributed timescales of reservoir dynamics on the long-term prediction ability.

\begin{figure}[b]
\includegraphics[width=0.95\linewidth]{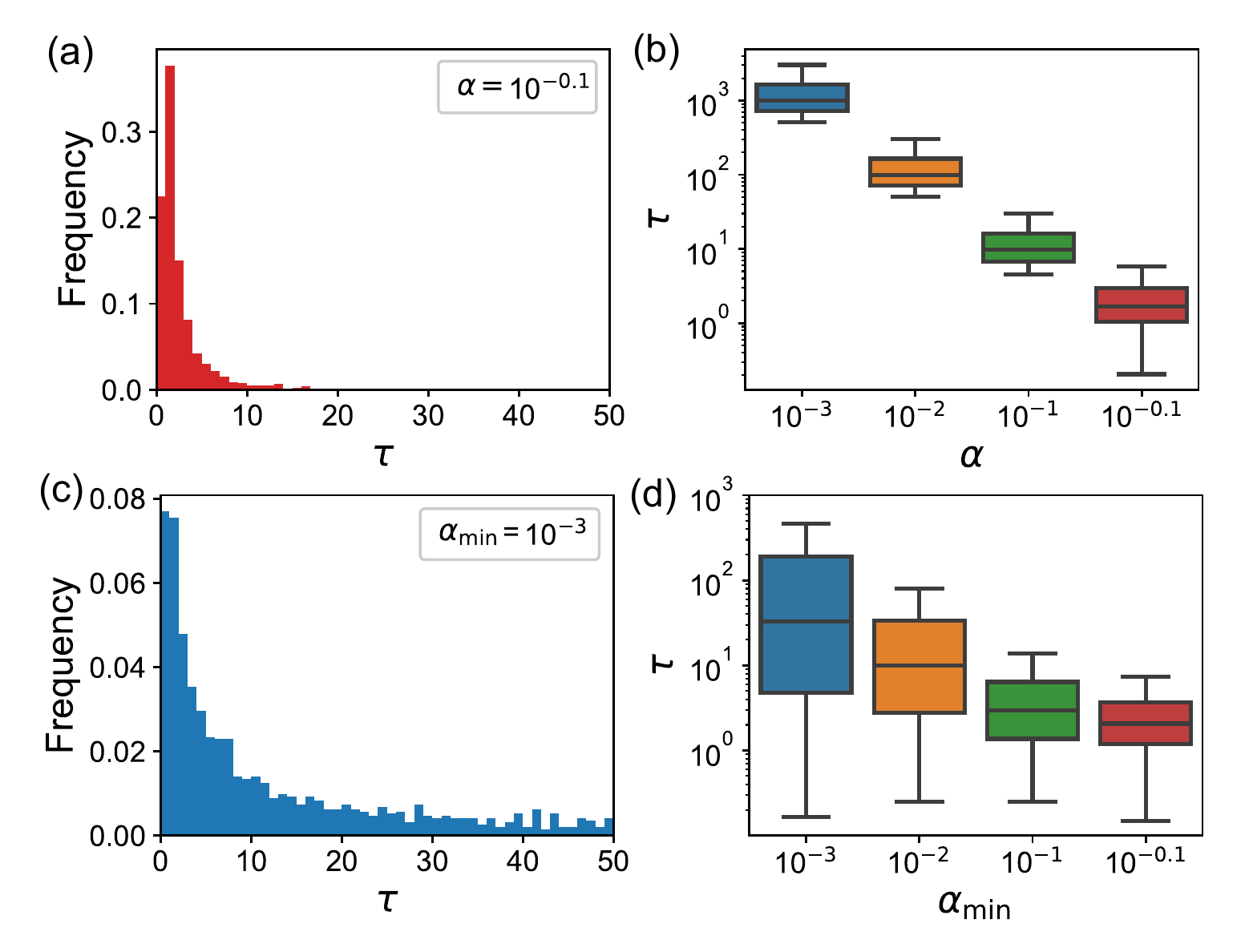}
\caption{\label{fig:timescale} Timescale distributions of reservoir dynamics in the linearized systems of the LI-ESN and the DTS-ESN  when $N_x=2000$ and $d=0.1$. (a) A histogram of timescales ($\tau_i$ in Eq.~(\ref{eq:tau_i})) in the LI-ESN with $\alpha=10^{-0.1}$. (b) The effect of $\alpha$ on the timescale distribution in the LI-ESN. (c) The same as (a) but in the DTS-ESN with $\alpha_{\rm max}=1$ and $\alpha_{\rm min}=10^{-3}$. (d) The effect of $\alpha_{\rm min}$ on the timescale distribution in the DTS-ESN with $\alpha_{\rm max}=1$.}
\end{figure}

\begin{table*}[t]
 \caption{\label{tab:ds}Chaotic fast-slow dynamical systems models used for prediction tasks. Each model consists of fast and slow subsystems of the variables specified in the third and fourth columns. The parameter values were set to exhibit chaotic dynamics.}
 \begin{ruledtabular}
\begin{tabular}{l|l|l|l|l}
\textrm{Model}& \textrm{Equations}& \textrm{Fast} & \textrm{Slow} & \textrm{Parameter values}\\
\colrule
(i) Rulkov \cite{rulkov2001regularization} & $x_{t+1}=\eta/(1+x^2_t)+y_t,~y_{t+1}= y_t-\mu (x_t-\sigma)$ & $x$ & $y$ & $\eta=4.2$, $\mu=0.001$, $\sigma=-1.0$ \\
\hline
(ii) \raisebox{-0.2em}{Hindmarsh-Rose \cite{hindmarsh1984model}} & \raisebox{-0.2em}{$\dot{x}=y-x^3+bx^2-z+I,~\dot{y}=1-5x^2-y,$} & \raisebox{-0.2em}{$x$, $y$} & \raisebox{-0.2em}{$z$} & \raisebox{-0.2em}{$b=3.05$, $I=3$, $\epsilon=0.01$,}\\
& $\dot{z}=\epsilon(s(x-x_0)-z)$ & & & $s=4$, $x_0=-1.6$ \\
\hline
(iii) \raisebox{-0.2em}{tc-VdP \cite{champion2019discovery}} & \raisebox{-0.2em}{$\dot{x}_1 = (y_1+c_1x_2)/\tau_1,~\dot{y}_1=(\mu_1(1-x_1^2)y_1-x_1)/\tau_1$} & \raisebox{-0.2em}{$x_1$, $y_1$} & \raisebox{-0.2em}{$x_2$, $y_2$} & \raisebox{-0.2em}{$\mu_1=5$, $\tau_1=0.1$, $c_1=0.001$} \\
& $\dot{x}_2 = (y_2+c_2x_1)/\tau_2,~\dot{y}_2=(\mu_2(1-x_2^2)y_2-x_2)/\tau_2$ & & & $\mu_2=5$, $\tau_2=1$, $c_2=1$ \\
\hline
(iv) \raisebox{-0.2em}{tc-Lorenz \cite{boffetta1998extension}} & \raisebox{-0.2em}{$\dot{X}=a(Y-X)$,~$\dot{Y}=r_sX-ZX-Y-\epsilon_sxy$,} & \raisebox{-0.2em}{$x$, $y$, $z$} & \raisebox{-0.2em}{$X$, $Y$, $Z$} & \raisebox{-0.2em}{$a=10$, $b=8/3$, $c=10$} \\
& $\dot{Z}=XY-bZ$,~$\dot{x}=ca(y-x)$, & & & $r_s=28$, $r_f=45$, \\
 & $\dot{y}=c(r_fx-zx-y)+\epsilon_f Yx$,~$\dot{z}=c(xy-bz)$ & & & $\epsilon_s=0.01$, $\epsilon_f=10$
\end{tabular}
\end{ruledtabular}
\end{table*}

The DTS-ESN is formulated as follows:
\begin{eqnarray}
  \mathbf{x}(t+\Delta t) &=& (I-A) \mathbf{x}(t) + A f(\mathbf{h}(t)), \label{eq:x}\\
  \mathbf{h}(t) &=& \rho W \mathbf{x}(t)+\gamma W^{\rm in}\mathbf{u}(t+\Delta t)+\zeta W^{\rm fb} \mathbf{y}(t), \label{eq:h}\\
  \mathbf{y}(t+\Delta t) &=&
  \left\{
  \begin{array}{l}
    \hspace{-1mm} W^{\rm out} [\mathbf{x}(t+\Delta t); \mathbf{u}(t+\Delta t); 1]~({\rm Task~1}) \\
    \hspace{-1mm} W^{\rm out} \mathbf{x}(t+\Delta t)~({\rm Task~2}),
  \end{array}
  \right.
    \label{eq:y}
\end{eqnarray}
where $t$ is the time, $\Delta t$ is the time step, $\mathbf{x}(t) \in \mathbb{R}^{N_x}$ is the reservoir state vector, $I \in \mathbb{R}^{N_x\times N_x}$ is the identity matrix, $A={\rm diag}(\alpha_1,\ldots,\alpha_{N_x})$ is the diagonal matrix of leak rates $\alpha_i$ for $i=1,\ldots,N_x$, $\mathbf{h}(t)$ is the internal state vector, $f$ is the element-wise activation function given as $f=\tanh$ in this study, $\rho W \in \mathbb{R}^{N_x\times N_x}$ is the reservoir weight matrix with spectral radius $\rho$, $\gamma W^{\rm in} \in \mathbb{R}^{N_x \times N_u}$ is the input weight matrix with input scaling factor $\gamma$, $\mathbf{u}(t)\in \mathbb{R}^{N_u}$ is the input vector, $\zeta W^{\rm fb} \in \mathbb{R}^{N_x \times N_y}$ is the feedback weight matrix with feedback scaling factor $\zeta$, $\mathbf{y}(t) \in \mathbb{R}^{N_y}$ is the output vector, and $W^{\rm out}$ is the output weight matrix. In the readout, we concatenate the reservoir state, the input, and the bias for Task 1 and use only the reservoir state for Task 2 as described in Eq.~(\ref{eq:y}). Only $W^{\rm out}$ is trainable and all the other parameters are fixed in advance \cite{jaeger2001echo, lukovsevivcius2009reservoir}. The DTS-ESN is reduced to the LI-ESN \cite{jaeger2007optimization} if $\alpha_i=\alpha$ for all $i$.

The fraction of non-zero elements in $W$ was fixed at $d=0.1$, which were randomly drawn from a uniform distriubtion in [-1,1], and $W$ was rescaled so that its spectral radius equals 1. The entries of $W^{\rm in}$ and $W^{\rm fb}$ were randomly drawn from a uniform distribution in $[-1,1]$. The leak rates were assumed to follow a reciprocal (or log-uniform) distribution in $[\alpha_{\rm min},\alpha_{\rm max}]$. This means that $\log_{10}\alpha_i$ was randomly drawn from a uniform distribution in $[\log_{10}\alpha_{\rm min}, \log_{10}\alpha_{\rm max}]$. The leak rate $\alpha_i$ of neuron $i$ is represented as $\alpha_i=\Delta t/c_i$ where $c_i$ denotes the time constant in the corresponding continuous-time model (see the Supplemental Material \cite{supplementary}).

In the training phase, an optimal output weight matrix $\hat{W}^{\rm out}$ is obtained by minimizing the following sum of squared output errors plus the regularization term:
\begin{eqnarray}
  &&\sum_{k} ||\mathbf{y}(t+k\Delta t)-\mathbf{d}(t+k\Delta t) ||_2^2 + \beta ||W^{\rm out}||_F^2, \label{eq:error}
\end{eqnarray}
where the summation is taken for all discrete time points in the training period, $||\cdot ||_2$ indicates the L2 norm, $||\cdot||_F$ indicates the Frobenius norm, $\mathbf{d}(t)$ denotes the target dynamics, and $\beta$ represents the regularization factor \cite{lukovsevivcius2009reservoir}. In the testing phase, $\hat{W}^{\rm out}$ is used to produce predicted outputs.

\begin{table}[b]
 \caption{\label{tab:datalen}The time steps and the durations of transient, training, and testing periods.}
 \begin{ruledtabular}
   \begin{tabular}{l|r|r|r|r}
\textrm{Model}& $\Delta t$ & $T_{\rm trans}$ & $T_{\rm train}$ & $T_{\rm test}$\\
\colrule
(i) Rulkov & 1 & 4000 & 8000 & 4000 \\
(ii) Hindmarsh-Rose & 0.05 & 200 & 1200 & 600 \\
(iii) tc-VdP & 0.01 & 50 & 150 & 100 \\
(iv) tc-Lorenz & 0.01 & 30 & 60 & 30
   \end{tabular}
 \end{ruledtabular}
\end{table}

{\it Analyses.--} The timescales of the reservoir dynamics in the DTS-ESN are mainly determined by the hyperparameters including the time step $\Delta t$, the spectral radius $\rho$, and the leak rate matrix $A$. The timescales of reservoir dynamics in the linearized system of Eq.~(\ref{eq:x}), denoted by $\tau_i$ for $i=1,\ldots,N_x$, are linked to the set of eigenvalues $\lambda_i$ of its Jacobian matrix as follows (see the Supplemental Material \cite{supplementary}):
\begin{eqnarray}
  \tau_i &=& -\frac{\Delta t}{\ln |\lambda_i|}. \label{eq:tau_i}
\end{eqnarray}

Figure~\ref{fig:timescale} demonstrates that the timescale distribution in the linearized system is different between the LI-ESN and the DTS-ESN. Figure~\ref{fig:timescale}(a) shows a timescale distribution for the LI-ESN with $\alpha=10^{-0.1}$. When $\alpha$ is decreased to produce slower dynamics, the peak of the timescale distribution increases while the distribution range in the logarithmic scale is almost unaffected as shown in Fig.~\ref{fig:timescale}(b) \cite{manneschi2021exploiting}. Figure~\ref{fig:timescale}(c) shows a broader timescale distribution of the DTS-ESN with $[\alpha_{\rm min},\alpha_{\rm max}]=[10^{-3},1]$. As shown in Fig.~\ref{fig:timescale}(d), the distribution range monotonically increases with a decrease in $\alpha_{\rm min}$ (see the Supplemental Material \cite{supplementary}). The timescale analysis based on the linearized systems indicates that the DTS-ESN with a sufficiently small $\alpha_{\rm min}$ has much more diverse timescales than the LI-ESN.

\begin{figure}[t]
\includegraphics[width=0.9\linewidth]{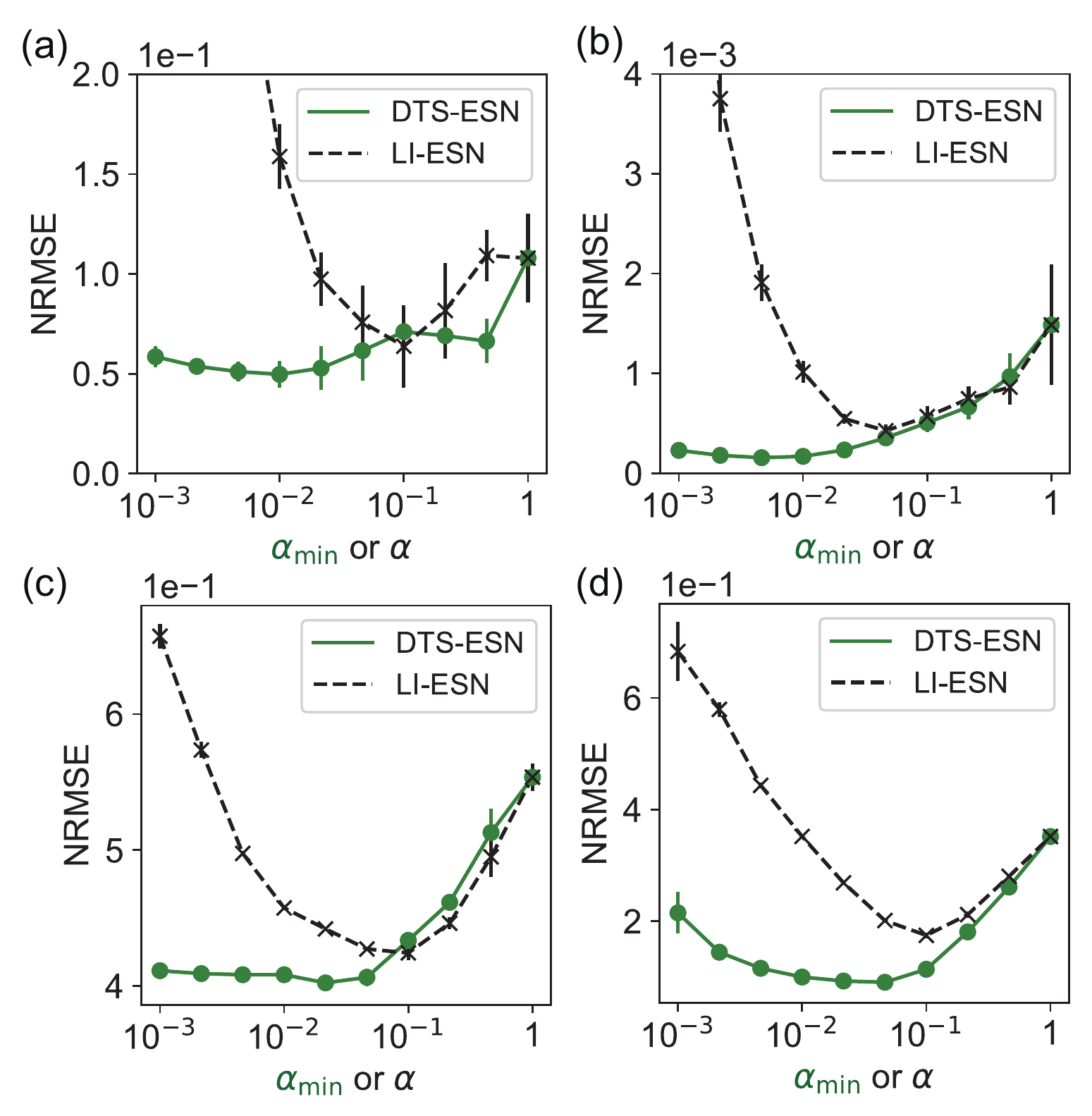}
\caption{\label{fig:nrmse} Comparisons of the testing errors (NRMSEs) between the DTS-ESN with $\alpha_i\in [\alpha_{\rm min},1]$ (green circles) and the LI-ESN with $\alpha_i=\alpha$ (black crosses). The marks indicate the average errors over ten simulation runs with different reservoir realizations. The error bar indicates the variance. The common parameter values are $\zeta=0$ and $\beta=10^{-3}$. (a) The Rulkov model. $N_x=200$ and $\gamma=\rho=1$. (b) The HR model. $N_x=200$ and $\gamma=\rho=1$. (c) The tc-VdP model. $N_x=200$ and $\gamma=\rho=1$. (d) The tc-Lorenz model. $N_x=400$ and $\gamma=\rho=0.1$.}
\end{figure}

\begin{figure}[t]
\includegraphics[width=\linewidth]{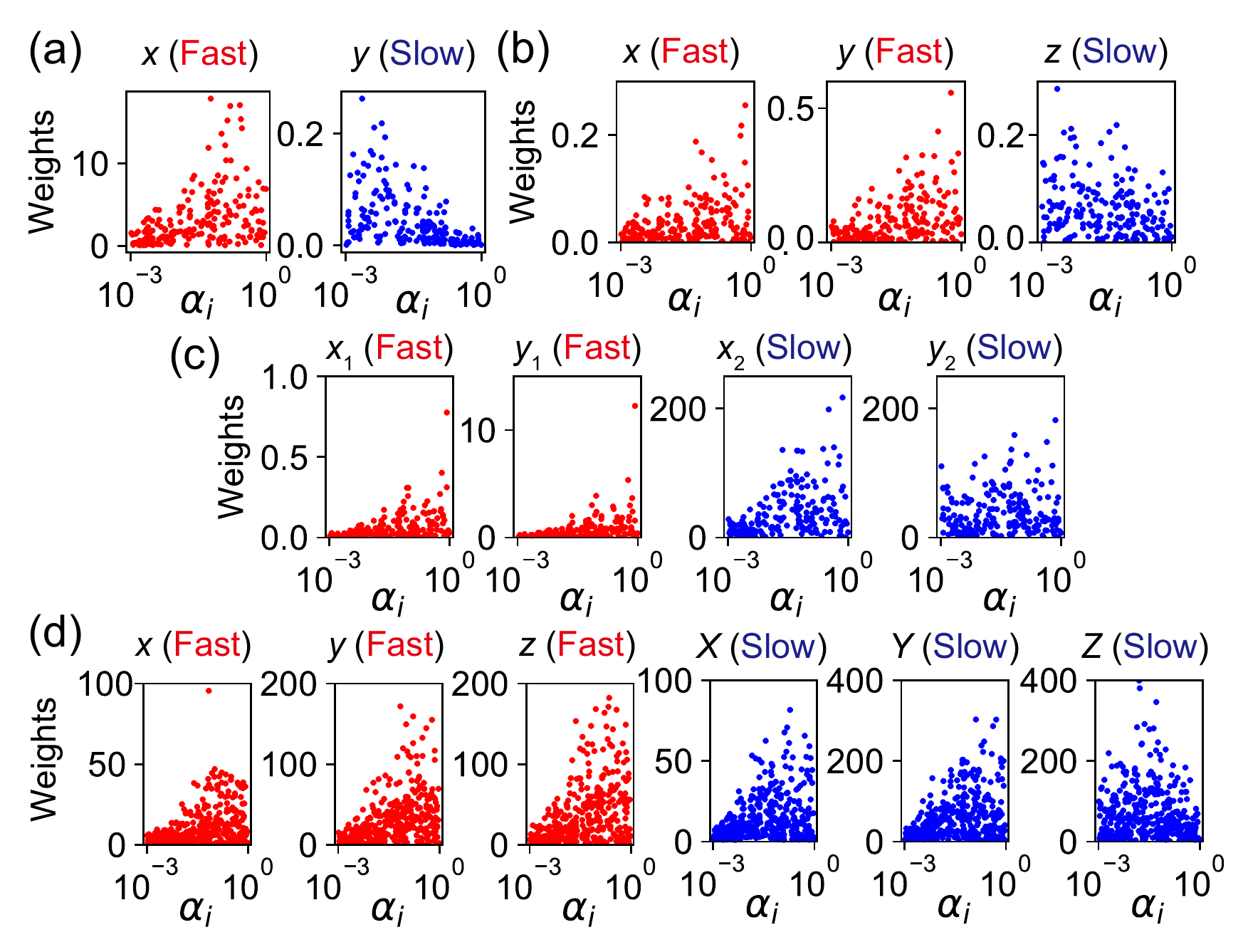}
\caption{\label{fig:Wout} The absolute output weights in $\hat{W}^{\rm out}$ of trained DTS-ESNs, plotted against the leak rates of the corresponding reservoir neurons. Each panel corresponds to each subsystem. The parameter values are the same as those for Fig.~\ref{fig:nrmse} but with $\alpha_{\rm min}=10^{-3}$. (a) The Rulkov model. (b) The HR model. (c) The tc-VdP model. (d) The tc-Lorenz model.}
\end{figure}

\begin{figure}[t]
\includegraphics[width=0.85\linewidth]{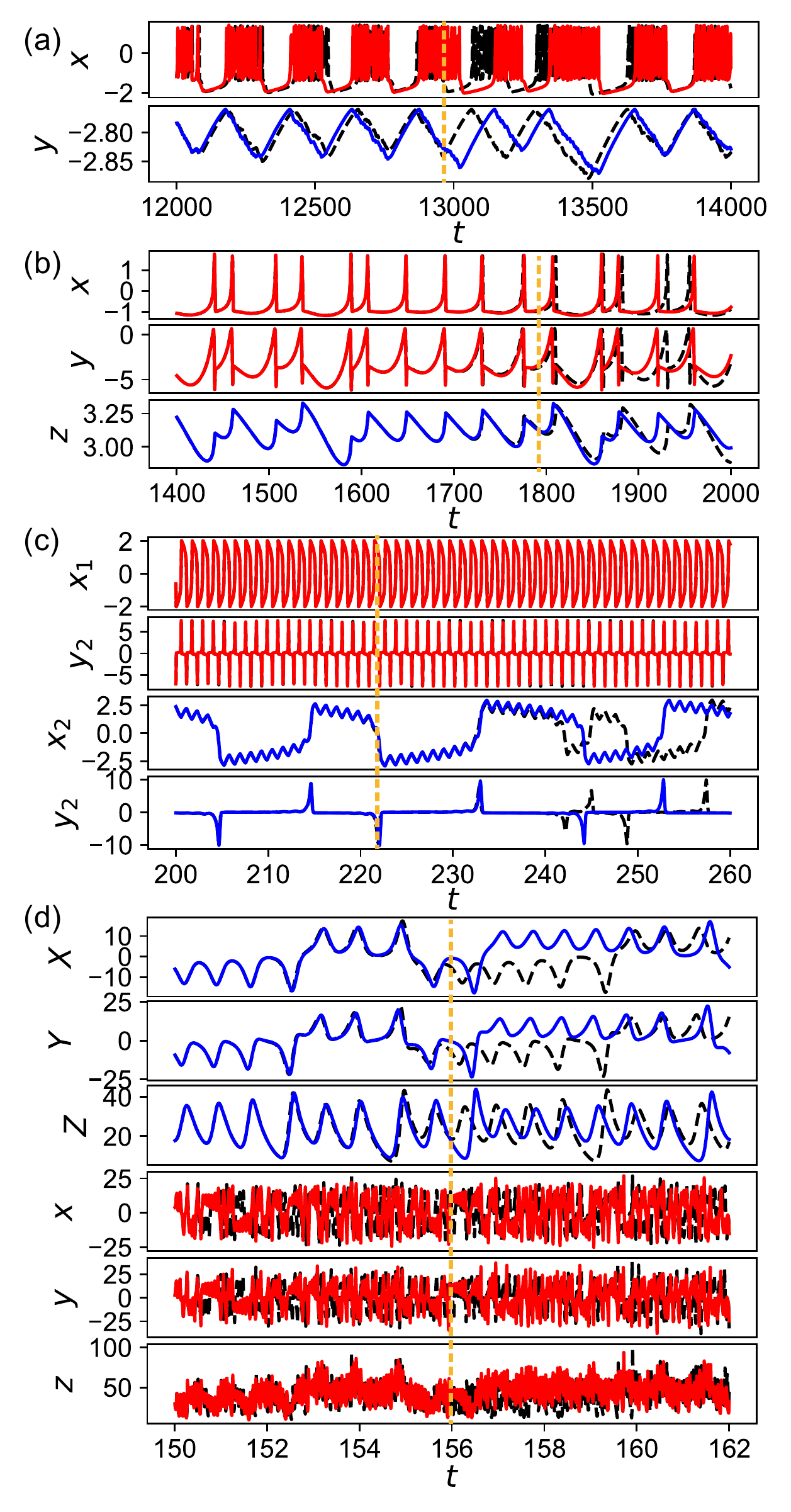}
\caption{\label{fig:closedloop} Examples of predictions (red and blue solid lines) by the closed-loop DTS-ESN models, superimposed on the target time series (black dashed lines). The orange vertical dashed line indicates the valid time. (a) The Rulkov model. $N_x=400$, $\gamma=0.8$, $\rho=1$, $\zeta=1$, $\beta=10^{-3}$, $\alpha_{\rm min}=10^{-6/9}$, and $\epsilon=0.01$. (b) The HR model. $N_x=400$, $\gamma=0.6$, $\rho=0.2$, $\zeta=0.4$, $\beta=10^{-3}$, $\alpha_{\rm min}=10^{-24/9}$, and $\epsilon=0.05$. (c) The tc-VdP model. $N_x=400$, $\gamma=0.1$, $\rho=0.03$, $\zeta=0.2$, $\beta=10^{-6}$, $\alpha_{\rm min}=10^{-2/9}$, and $\epsilon=0.4$. (d) The tc-Lorenz model. $N_x=1000$, $\gamma=0.01$, $\rho=0.01$, $\zeta=0.04$, $\beta=10^{-4}$, $\alpha_{\rm min}=10^{-8/9}$, and $\epsilon=0.4$.}
\end{figure}

\begin{figure}[t]
\includegraphics[width=0.9\linewidth]{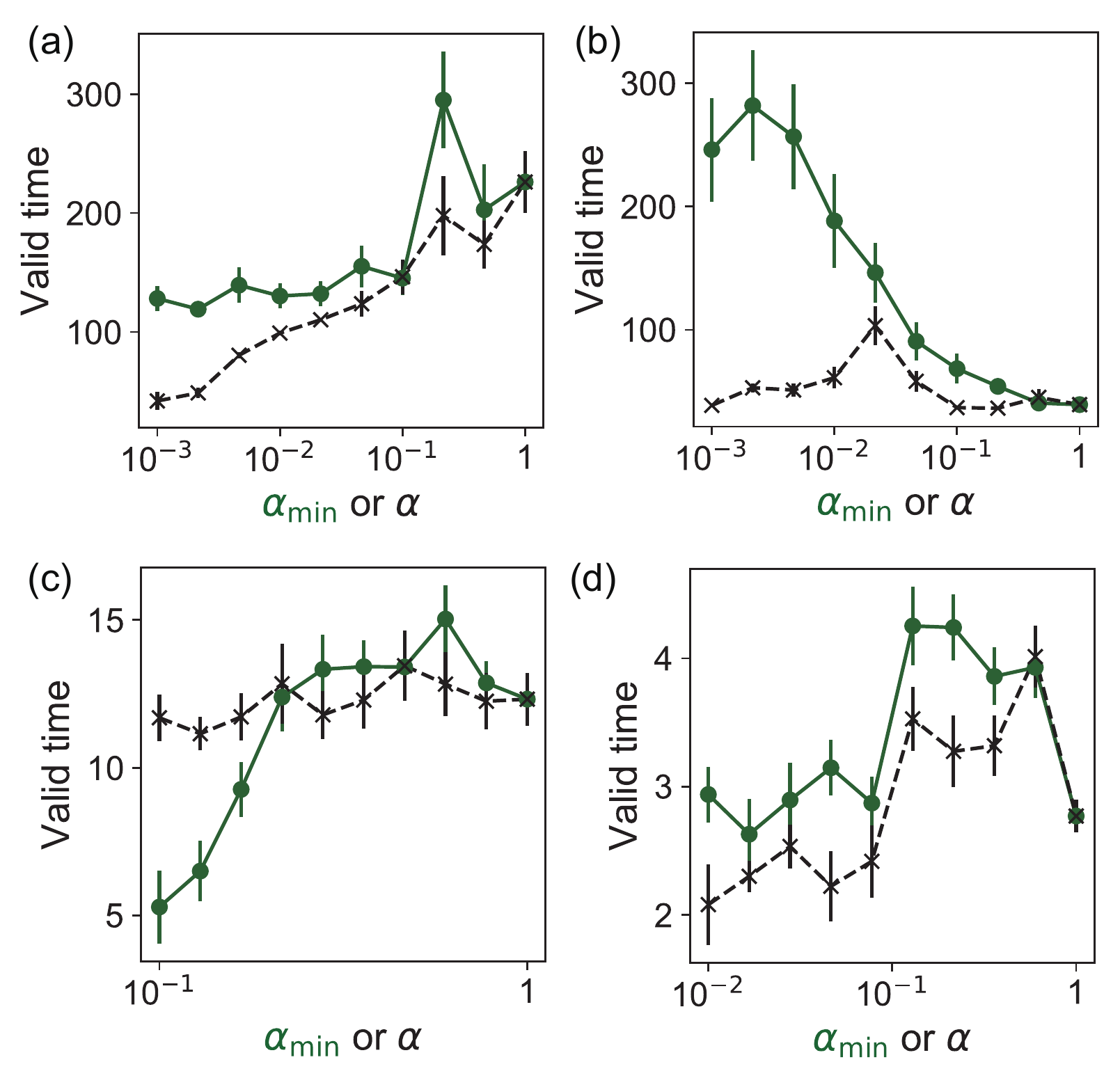}
\caption{\label{fig:validtime} Comparisons of the valid time between the DTS-ESN (green circles) and the LI-ESN (black crosses). The marks indicate the average values over 20 simulation runs with different reservoir realizations. The error bar indicates the standard error. The parameter conditions and data settings are the same as those for Fig.~\ref{fig:closedloop}, except for the varied one. (a) The Rulkov model. (b) The Hindmarsh-Rose model. (c) The tc-VdP model. (d) The tc-Lorenz model.}
\end{figure}


{\it Results.--} We evaluated the computational performance of the DTS-ESN in prediction tasks involved with four chaotic fast-slow dynamical systems models listed in Table~\ref{tab:ds}: (i) the Rulkov model which is a 2D map replicating chaotic bursts of neurons \cite{rulkov2001regularization}; (ii) the Hindmarsh-Rose (HR) model which is a phenomenological neuron model exhibiting irregular spiking-bursting behavior \cite{hindmarsh1984model}; (iii) the two coupled Van der Pol (tc-VdP) model which is a combination of fast and slow limit cycle oscillators with nonlinear damping \cite{champion2019discovery}; (iv) the two coupled Lorenz (tc-Lorenz) model which is a caricature representing the interaction of the ocean with slow dynamics and the atmosphere with fast dynamics \cite{boffetta1998extension,borra2020effective}. There is a large timescale gap between the fast and slow subsystems. 

We generated time series data from each dynamical system model using the parameter values listed in Table~\ref{tab:ds}. The ODE models were numerically integrated using the Runge-Kutta method with time step $\Delta t$. Then, we separated the whole time series data of total length $T_{\rm total}$ into transient data of length $T_{\rm trans}$, training data of length $T_{\rm train}$, and testing data of length $T_{\rm test}$. The transient data was discarded to wash out the influence of the initial condition of the reservoir state vector. We set the time step and the durations of data as listed in Table~\ref{tab:datalen} unless otherwise noted.
  
First, we performed the one-step-ahead prediction task (Task 1) using the open-loop model shown in Fig.~\ref{fig:DTSESN}(b), where the states of the whole variables at one step ahead are predicted only from the input time series of fast variables. The prediction performance is evaluated with the following Normalized Root Mean Squared Error (NRMSE) between the model predictions and the target outputs:
\begin{eqnarray}
  \mbox{NRMSE} &=& \frac{\sqrt{\langle ||\mathbf{y}(t+k\Delta t)-\mathbf{d}(t+k\Delta t)||_2^2 \rangle_k}}{\sqrt{\langle ||\mathbf{d}(t+k\Delta t)-\langle \mathbf{d}(t+k\Delta t)\rangle_k||^2_2 \rangle_k}}, \label{eq:NRMSE}
\end{eqnarray}
where $\langle \cdot \rangle_k$ denotes an average over the testing period.



Figures~\ref{fig:nrmse}(a)-(d) show the comparisons between the DTS-ESN and the LI-ESN in the NRMSEs for the four dynamical systems models listed in Table~\ref{tab:ds} (see the Supplemental Material for examples of predicted time series \cite{supplementary}). The horizontal axis is $\alpha_{\rm min}$ for the DTS-ESN and $\alpha$ for the LI-ESN. When $\alpha_{\rm min}=\alpha=1$, the two models coincide and yield the same performance. The prediction performance is improved as $\alpha_{\rm min}$ is decreased from 1 for the DTS-ESN, mainly due to an increase in the prediction accuracy with respect to the slow variables (see the Supplemental Material \cite{supplementary}). The DTS-ESN can keep a relatively small prediction error when $\alpha_{\rm min}$ is decreased even to $10^{-3}$ in contrast to the LI-ESN. The best performance of the DTS-ESN is obviously better than that of the LI-ESN for all the target dynamical systems, indicating a higher ability of the DTS-ESN. Moreover, even with $\alpha_{\rm min}$ fixed at $10^{-3}$, the DTS-ESN can achieve the performance comparable to the best one obtained by the LI-ESN for all the target dynamics. This effortless setting of leak rates is an advantage of the DTS-ESN over the LI-ESN and the hierarchical LI-ESNs.

Figures~\ref{fig:Wout}(a)-(d) demonstrate the absolute output weights in $\hat{W}^{\rm out}$ of the trained DTS-ESNs, plotted against the leak rates of the corresponding neurons. Each panel corresponds to an output neuron for the variable specified by the label on that. The results indicate that the reservoir neurons with large $\alpha_i$, having small timescales, are mainly used for approximating the fast subsystems (red points) and those with small $\alpha_i$, having large timescales, are for the slow subsystems (blue points). We can see that the neuronal states with appropriate timescales are selected to comply with the timescale of the desired output as a result of model training. In Fig.~\ref{fig:Wout}(d) for the tc-Lorenz model, the reservoir neurons with relatively small $\alpha_i$ are used for the slow variable $Z$ but not for the other slow variables ($X$ and $Y$). This means that the dynamics of $X$ and $Y$ are essentially not as slow as that of $Z$, causing the performance degradation with a large decrease in $\alpha_{\rm min}$ as shown in Fig.~\ref{fig:nrmse}(d). By increasing the reservoir size and the length of training data, this degradation is mitigated (see Supplemental Material \cite{supplementary}). The natural separation of the roles of neurons can be regarded as a spontaneous emergence of modularization found in many biological systems \cite{kashtan2005spontaneous,lorenz2011emergence,yang2019task}.

  Second, we performed the autoregressive prediction task (Task 2). As shown in Fig.~\ref{fig:DTSESN}(c), the open-loop model is trained using the training data, and then the closed-loop model is used to generate predicted time series autonomously. We increased $T_{\rm train}$ from 60 in Table~\ref{tab:datalen} to 120 for the tc-Lorenz model, in order to improve the prediction performance. In the testing phase, we evaluated the valid time \cite{pathak2018model} for the slow dynamics, indicating the elapsed time duration (measured in actual time units) before the normalized prediction error $E(t)$ exceeds a threshold value $\epsilon$ in the testing phase. The error $E(t)$ is defined as follows \cite{pathak2018hybrid}:
\begin{eqnarray}
  E(t) &=& \frac{||\tilde{\mathbf{x}}(t)-\tilde{\mathbf{d}}(t)||_2}{\langle ||\tilde{\mathbf{d}}(t+k\Delta t)||_2^2 \rangle_k^{1/2}}, \label{eq:VT}
\end{eqnarray}
where $\tilde{\mathbf{x}}(t)$ and $\tilde{\mathbf{d}}(t)$ represent the reservoir state vector and the target vector of slow variables, respectively, and $\langle \cdot \rangle_k$ denotes an average over the testing period.

Figures~\ref{fig:closedloop}(a)-(d) show examples of autoregressive predictions (red lines for fast variables and blue lines for slow ones) by the closed-loop DTS-ESN models for the four target dynamical systems. The predicted time series approximates well the target time series (black dashed lines) until the valid time indicated by the vertical dashed line. The divergence of the prediction error after a finite time is inevitable due to the chaotic dynamics. We note that, in Fig.~\ref{fig:closedloop}(c) for the tc-VdP model, the discrepancy between the predicted and target time courses is not prominent until around $t=240$ but the normalized error exceeds $\epsilon=0.4$ at around $t=221$. If the threshold value is changed to $\epsilon=3$, the valid time is increased to around 44. Figures~\ref{fig:validtime}(a)-(d) demonstrate the comparisons of the valid time between the closed-loop DTS-ESN and the closed-loop LI-ESN for the four target dynamical systems. In all the panels, the largest valid time is achieved by the DTS-ESN, suggesting its higher potential in the long-term prediction. The valid time largely depends on the hyperparameter setting including the range of leak rates, which may be associated with the attractor replication ability of the DTS-ESN as measured by the Lyapunov exponents (see Supplemental Material \cite{supplementary} for details).

{\it Discussion and conclusion.--} We have proposed the RC model with diverse timescales, the DTS-ESN, by incorporating distributed leak rates into the reservoir neurons for modeling and prediction of multiscale dynamics. The results of the prediction tasks indicate the effectiveness of our randomization approach to realizing a reservoir with rich timescales. Although we assumed a specific distribution of leak rates in this study, another distribution could further improve the prediction performance. Moreover, another type of heterogeneity of reservoir components could boost the ability of RC systems in approximating a wide variety of dynamical systems. Future applications include a prediction of large-scale spatiotemporal chaotic systems \cite{pathak2018model,vlachas2020backpropagation} from partial observations and an inference of slow dynamics from experimentally measured data involved in real-world multiscale systems \cite{meinen2020observed,proix2018predicting}.

{\it Acknowledgmnents.--} We thank Daisuke Inoue for fruitful discussion and Huanfei Ma for valuable comments. This work was performed as a project of Intelligent Mobility Society Design, Social Cooperation Program (Next Generation Artificial Intelligence Research Center, the University of Tokyo with Toyota Central R\&D Labs., Inc.), and supported in part by JSPS KAKENHI Grant Numbers 20K11882 (GT), JP20H05921 (KA), JST-Mirai Program Grant Number JPMJMI19B1 (GT), JST CREST Grant Number JPMJCR19K2 (GT), JST Moonshot R\&D Grant Number JPMJMS2021 (KA), AMED under Grant Number JP21dm0307009 (KA), and International Research Center for Neurointelligence at The University of Tokyo Institutes for Advanced Study (UTIAS) (GT, KA).

\bibliographystyle{apsrev4-2}

%

\end{document}